\documentclass[a4paper, 10pt, conference]{IEEEtran}
\usepackage{amsmath,amssymb,amsfonts}
\usepackage{algorithmic}
\usepackage{graphicx}
\usepackage{textcomp}
\usepackage{xcolor}
\usepackage[hidelinks]{hyperref}
\usepackage{xurl}
\usepackage[sort, numbers]{natbib}
\usepackage[user,titleref]{zref}

\renewcommand\harvardurl[1]{\textbf{URL:} \url{#1}}

\def\BibTeX{{\rm B\kern-.05em{\sc i\kern-.025em b}\kern-.08em
    T\kern-.1667em\lower.7ex\hbox{E}\kern-.125emX}}
\begin{document}

\title{Why Report Failed Interactions With Robots?!\\ Towards Vignette-based Interaction Quality}
\author{
\IEEEauthorblockN{Agnes Axelsson}
\IEEEauthorblockA{\textit{TU Delft} \\
Delft, Netherlands \\
a.axelsson@tudelft.nl}
\and
\IEEEauthorblockN{Merle Reimann}
\IEEEauthorblockA{\textit{VU Amsterdam} \\
Amsterdam, Netherlands \\
m.m.reimann@vu.nl}
\and
\IEEEauthorblockN{Ronald Cumbal}
\IEEEauthorblockA{\textit{Uppsala University} \\
Uppsala, Sweden \\
ronald.cumbal@it.uu.se}
\and
\IEEEauthorblockN{Hannah Pelikan}
\IEEEauthorblockA{\textit{Linköping University} \\
Linköping, Sweden \\
hannah.pelikan@liu.se}
\and
\IEEEauthorblockN{Divesh Lala}
\IEEEauthorblockA{\textit{Osaka University} \\
Osaka, Japan \\
lala.divesh.kanu.es@\\osaka-u.ac.jp}
}

\maketitle

\begin{abstract}
Although the quality of human-robot interactions has improved with the advent of LLMs, there are still various factors that cause systems to be sub-optimal when compared to human-human interactions. The nature and criticality of failures are often dependent on the context of the interaction and so cannot be generalized across the wide range of scenarios and experiments which have been implemented in HRI research. In this work we propose the use of a technique overlooked in the field of HRI, ethnographic vignettes, to clearly highlight these failures, particularly those that are rarely documented. We describe the methodology behind the process of writing vignettes and create our own based on our personal experiences with failures in HRI systems. We emphasize the strength of vignettes as the ability to communicate failures from a multi-disciplinary perspective, promote transparency about the capabilities of robots, and document unexpected behaviours which would otherwise be omitted from research reports. We encourage the use of vignettes to augment existing interaction evaluation methods.
\end{abstract}


\section{Introduction}

High-quality dialogue with robots is a goal for many human-robot interaction (HRI) researchers \cite{kennington2024dialoguerobotsproposalsbroadening}. Despite technological advancements, dialogues in HRI sometimes fail. In this paper, we propose vignette-writing as a method for reporting observations from failed interactions.

The abilities of large language models (LLMs) to simulate human language have sparked an increased interest and optimism towards generating meaningful dialogues, despite their well-known shortcomings \cite{bommasani2023foundation, bender2021dangers,elazar2021measuring}. 
However, there is still much ground to cover towards flawless spoken interactions with robots \cite{marge2022spoken}.
One of the challenges that need to be addressed in order to move towards this goal lies in \textbf{defining, describing and evaluating} concrete interactions. In this paper, we propose that describing moments of failure in dialogues through \textbf{ethnographic} methods is one path to understanding, evaluating and defining human-robot interactions. We draw on the long history of human-machine dialogue in human-computer interaction (HCI) \cite{Luff_Gilbert_Frohlich_1990,weizenbaum1966eliza,Porcheron_Fischer_Reeves_Sharples_2018} and on work on speech-enabled robots \cite{marge2022spoken, skantze2021turntakinglitreview,reimann2024survey}.

Analysing human-agent\footnote{For the purposes of this paper, an \textit{agent} is a virtual or situated agent, while a \textit{robot} is strictly a situated embodied agent.} spoken interactions is challenging. Unconstrained interactions will always lead to unexpected edge cases which can affect the experience of the user in ways that were not anticipated by the system designers. The users' expectations of the robot's capabilities, informed by embodiment and multimodal capabilities, can also be unpredictable. This leads to varying expectations of interaction quality \cite{kunold2023not}. More importantly, evaluating interaction typically involves qualitative \cite{shin2022usercenteredexploration, berzuk2022morethanwordsframeworkfordescribingdialogdesigns} or quantitative \cite{kim2024understandingllmpoweredhri, dehaas2024wokebot} measurements that may only capture specific aspects of the interaction, potentially overlooking subtly important factors. Given this, we need to reexamine the analysis and reporting of spoken HAI.

\textbf{Failures}, a ``degraded state of ability which causes the behaviour or service being performed by the system to deviate from the ideal, normal or correct functionality'' \cite{somasundaram2023intelligentdisobedience, brooks2017human}, are a particularly interesting phenomenon in HAI evaluation. Interactive agents can employ remediation strategies \cite{Jefferson_2018} to repair the consequences if they are aware of a failure that is not too big to excuse or explain \cite{correia2018exploring}. Which strategies are appropriate depends on the nature of the interaction \citep{chang2024sorrytokeepyouwaiting}.
Failures consistently lead to human users losing trust in robots \cite{desai2012effects,rossi2017timing}. Users may lose trust in their own ability to contribute to the interaction, or lose trust in the interaction itself \cite{rudaz2023mpcosin}.

In this paper, we propose the use of ethnographic vignettes \cite{Bloom-Christen_Grunow_2022, emerson2011writing} -- short descriptions of situated, contextualised interactions -- to add to classic interaction assessment metrics by reporting moments of failure in interactions. Vignettes let us report unexpected events and deviations from planned scenarios in HAI. As there is no consensus on what constitutes a \textit{good} or \textit{poor} dialogue interaction in human-agent interaction \cite{deriu2021survey}, qualitative approaches such as these vignettes are valuable for capturing subjective observations during interactions. Such observations can then provide insights that other researchers in the community may also find worthy of further analysis. 

We demonstrate the use of ethnographic vignettes by drawing from our personal experiences in building and observing robots in interaction. We illustrate how vignette-based reporting can further raise awareness of challenges in HRI, allowing researchers to document issues that are easily overlooked or considered ``unworthy'' to report. From a \textit{design ethnography} perspective, our vignettes resemble stories told to reflect and influence the stage of designing a system  \cite{Crabtree_Rouncefield_Tolmie_2012}. We use vignette writing as a tool for documenting observations by practitioners, as used for instance by teachers in the educational sciences to reflect on classroom experiences \cite{BehmCross_2017,Cree_2012}. 

\section{Related work}

\subsection{Quality in Human Interactions}
\label{sec:functional-perspectives-hhi}
One perspective on evaluating interactions between humans is measuring how interaction partners achieve their communicative goals. Communicative behaviours that lead to the partners not achieving their joint goals are then poor interactions, while those that lead to the partners achieving their goals quickly and efficiently (i.e. with few repairs \cite{Jefferson_2018}) are indicative of a good interaction \cite{clark1996using}. 

With his maxims of quantity, quality, relation and manner in the \textit{cooperative principle}, \citet{grice1975logicandconversation} highlights that contributions should be brief, true, relevant to the interaction and understandable to the people in the current interaction. Likewise, \citeauthor{clark1996using} and \citeauthor{allwood1992feedback} argue that communication is a cooperative activity that is negotiated between both parties of an interaction, although the exact mechanisms through which this happens, as well as the independence of the speaker from the listener, are not fully agreed upon \cite{clark1996using, allwood1992feedback}. Evaluating an interaction as an outsider, hence, becomes inherently subjective, as the true validity of an interaction can only be analysed from the perspectives of the participants' personal and shared goals.

\subsection{Evaluating Human-Robot Interactions}
\label{sec:interaction-quality}
Unlike humans as described by \citet{grice1975logicandconversation}, interactive machines do not genuinely follow a cooperative dynamic or inherently adapt to the needs of the user \cite{miehling2024languagemodelsdialogueconversational}. Users can, however, easily adapt to the machine \cite{suchman2007human} -- or to what they believe the machine is doing \cite{clarkfischer2023depictions}.

Evaluating interactions between humans and agents is heavily influenced by the dialogue system's characteristics and the task associated with the interaction \citep{deriu2021survey}. Numerous scales and measures have been proposed for the evaluation of interactions with artificial social agents, although there is no standard evaluation procedure. Rather, evaluation methods vary significantly between studies, since each method is often task and context dependent \cite{steinfeld2006common, coronado2022evaluating}. \citet{fitrianie2019we} and \citet{bagchi2023towardsimprovedreplicability} have criticised the low reuse of questionnaires at the IVA and HRI conferences respectively.

Following the principles from Section~\ref{sec:functional-perspectives-hhi}, a user's experience of HRI can also be evaluated by measuring satisfaction \citep{kocabalil2018measuring, campos2018challenges, borsci2022chatbot}, frustration \cite{thomason2015learning}, or the user’s overall perception of the agent to gain insights into the interaction \cite{foster2012two,milhorat2019conversational}. Tools like the Godspeed questionnaire \cite{bartneck2009measurement} or the extensive 90-item Artificial-Social-Agent questionnaire \cite{fitrianie2022artificial} are often employed. Post-interaction semi-structured or open interviews \cite{pollmann2023entertainmentvsmanipulation} complement the more static questionnaires. 

Moving away from the users' subjective experiences, quality can be measured by the objective number of turns taken \cite{lee2010hybrid}, the interaction duration \cite{lison2015hybrid} or, for task-based interactions, the task success rate \cite{lee2010hybrid}. Objective metrics offer the advantage that the robot can be made to evaluate the quality of its own interaction in real-time \cite{mayima2022towards}. 

Both subjective and objective data are valuable for evaluating interaction quality. These measures allow researchers to compare interactions within studies or across different related works. However, they often only capture the constructs the researchers intended to measure, potentially overlooking many elements of the interaction that may have impacted it but were not expected. 

\subsection{Observing Human-Robot Interactions}

We have described ways of \textit{measuring} subjective and objective factors after an interaction, or objective factors during an interaction. There is less work on \textit{observing} and documenting the quality of an interaction. An early study by \citeauthor{sabanovic2006robots} focused on documenting HRI in real-world settings. The authors pushed for the observation of social robots in natural environments, demonstrating that this approach highlighted significant issues in social robot design and challenged initial design assumptions, ultimately enhancing the robot's interaction qualities \citep{sabanovic2006robots}.

Methodologies inspired by ethnographic\footnote{As ethnography is the study of society and humans, an ethnographic perspective on HRI means observing the robot as it fits into society.} research are useful for describing a full interaction, but can be misunderstood or incorrectly perceived as lacking scientific validity \cite{jacobs2020evaluating}. \citeauthor{mutlu2008robots} analyzed ethnographic data collected from an autonomous delivery robot in a hospital, revealing substantial differences in how individuals integrated the robot into their work \citep{mutlu2008robots}.
\citeauthor{forlizzi2004assistive} showed varied reactions among users interacting with cleaning robots in a home environment. The study revealed that social attribution to the robots encouraged faster adoption and reduced the stigma associated with being reliant on the technology \citep{forlizzi2004assistive}. \citeauthor{sabelli2011conversational} observed a conversational robot in an elderly care center over a period of 3.5 months, gaining a deeper understanding of how elderly individuals interacted with the robot. Interactions were framed in terms of the needs and activities of daily life \citep{sabelli2011conversational}. While these studies provide ways to interactions over longer periods of time through observation, other approaches such as ethnomethodological conversation analysis allow to analyze specific moments in an interaction with the help of video recordings, scrutinizing what designers may consider a failed interaction in fine detail \citep{tisserand11unraveling, pelikan2016nao,Stommel_deRijk_Boumans_2022, rudaz2023inanimate}.

Observational studies are often framed as exploratory or initial steps in longer research projects, but they can be valuable in \textit{all} research evaluating and analysing human-agent interactions. As demonstrated by these studies, the authors discovered unexpected results that significantly enhanced the overall understanding of the interactions. While extensive qualitative studies are important to understand the wider societal impact of emerging systems, in this paper we want to introduce a method that enables developers to formulate observations during prototyping, testing and demonstration.

\section{Methodology}
Even though there are calls for focusing on failures in human-robot interaction \cite{belhassein2022addressingjointactionchallenges, breazeal2004socialinteractionsinhri}, they are not as consistently reported as intended or designed behaviours in scientific literature. By engaging in vignette-writing, we want to put more focus on reporting moments in HRI in which interactional trouble emerges from the perspective of the user, or in which the interactions did not evolve as intended by design. This stands in the conversation analytic tradition of \textit{deviant case analysis} \cite{Schegloff_1968,Sidnell_Stivers_2012}, where deviations from the typical interaction pattern are scrutinized with particular attention. 
By looking in detail at cases that could be considered \textit{outliers}, we hope to learn more about the persisting problems that dialogue systems face in HRI. 

We present our observations in the form of \textit{ethnographic vignettes} \cite{emerson2011writing, Bloom-Christen_Grunow_2022} focusing on illustrations of cases where interactions fell short of their intended quality. Systematically working through negative examples allows us to reason about interaction quality in an inductive way. We do not claim that we have found all problems that may possibly occur, but rather wish to contribute to a critical discourse that may enable HRI researchers to make informed design decisions while also considering HRI-specific challenges and limitations. 

Generalization and quantity is not the concern for the illustrating examples; the fact that these cases happened provide evidence that they do occur \cite{Schegloff_1993}. We thoroughly discussed each of the vignettes in our diverse research team, ensuring that the vignettes are related to spoken HRI, and describe anecdotal, but real, actionable and generalisable observations.

\subsection{How to Write Vignettes}
\label{sec:how-to-write-vignettes}
Writing vignettes requires a different writing style and author's voice than writing other scientific texts. They describe what the author observed and experienced \cite{jacobsen2014vignettesofinterviews} and are always shaped by the author's perspective. By properly formulating vignettes, observations can be mapped to social worlds and practices that are more generally understood. We adapted Emerson's guide to writing ethnographic fieldnotes \cite{emerson2011writing} to equip people without prior  training in ethnography to produce systematic descriptions of their observations. Because vignettes are a personal expression of an event and the takeaways from that event, they \textbf{will look different from person to person} -- some of the vignettes in Section~\ref{sec:vignettes} illustrate this variation within the bounds described in this section. 

For vignettes, the \textbf{writing style} is characterized by more details, adjectives and adverbs, especially when starting to write. As \citet[p. 131]{Goffman_1989} put it in a talk on fieldwork, ``as loose as that adverbialized prose is, it's still a richer matrix to start from than stuff that gets reduced into a few words of `sensible sentences'. [...] to be scientific in this area, you've got to start by trusting yourself and writing as fully and as lushly as you can. [...] put yourself into situations that you write about so that later on you will see how to qualify what it is you've said. You say, `If felt that,', `my feeling was,' `I had a feeling that' --- that kind of thing. This is part of the self-discipline.'' The writing is supposed to be reflective, so first-person pronouns are encouraged. 

Writing vignettes is also an \textbf{iterative process}, it means revising each vignette over and over. It can be helpful to come back to some hours later to fill in missing elements or to remove details that are less essential for mirroring what happened. It is usually better to add too much detail in the first iterations and then gradually remove the parts that are not needed. Our informal guide to writing ethnographic vignettes for HRI is located in Appendix~\ref{sec:appendix-vignette-writing}.

\section{Vignettes}
\label{sec:vignettes}
We provide six vignettes that describe failures and reflect on the broader current challenges for human-robot dialogue that the vignettes point to.

\subsection{Challenges during the system development}
While working on a system, developers often test their systems themselves in an ad hoc manner. Formulating these observations in vignettes can be a valuable way to report on design avenues that were not pursued further, or to outline problems encountered during testing that will likely cause trouble during the real-world deployment. 

\subsubsection{Vignette 1} \textit{LLM-generated study misinformation:}
\zlabel{vignette:hallucinating}

We were building the robotic system to explore what purposes it could serve in public transit. In early testing, we explored how to generate answers to unexpected questions by passengers. We speculated that a LLM could be applied to quickly generate answers to such questions based on TTS input from the passengers' speech, reducing the time to generate responses since it would take an unreasonable amount of time for the Wizard of Oz who was teleoperating our robot to type out an answer to each unexpected question.

The LLM was prompted with the relevant background details of the role it was supposed to play in our experiment. In testing, when asked why it was in the vehicle, the LLM-driven robot consistently responded with made-up details about the experimental setup and the researchers behind the experiment. When asked who the researchers were, the system provided email addresses and names of individuals who did not exist, but whose affiliations and identities would have been plausible to the uninformed participants. Similarly, the LLM-powered robot kept going beyond its prompted information when it was asked what the experiment was about, and added statements about what research questions were being explored, although this had not been asked for and was not part of the information to which it had access.

\paragraph*{Reflection} The readiness with which the system was ready to say something that we could tell was untrue, that the passengers would not be able to tell was untrue, and that it itself had no way of verifying, surprised us. We had expected the LLM to fill the dead air with small talk or harmless filler, but instead its behaviour could have easily been harmful to our passengers' experience. We ended up having the Wizard type out these responses manually, with the downside of significant delays. This is a context where Wizard of Oz makes sense as a way for the designer to \textit{play along} with the user's ideas and in-situ brainstorming about what the system should be able to do. We prompted the LLM to not make up details about the experiment, but the LLM's role was explicitly to address situations where the question was unexpected and not something we could have a pre-written answer for. Addressing such problems as they appear would be an unending \textit{qualification problem} \cite{ginsberg1988qualificationproblem}.

\subsubsection{Vignette 2} \textit{Being unable to recognize the user's name:}
\zlabel{vignette:name-recognition}
For a lab opening, I was preparing a demo of a patient interview robot. The code was already finished and parts of it had been tested on their own with generic answers. However, during the final tests, I tried to answer with different answers and a number of international colleagues tested the system as well.
The robot started the interaction by introducing itself and then asking for the name of the user. When I or my colleagues provided our names, the robot was unable to understand them correctly and either misunderstood them as other words, or asked for a repetition. I tried to get the robot to understand my name by pronouncing it in various ways, but the robot's speech recognition failed repeatedly. While we first laughed and saw it as a challenge to get the robot to say our names, it soon became frustrating since we noticed that it was just not possible to use our real names in the conversation.

\paragraph*{Reflection} This vignette illustrates how even a task that appears simple can fail consistently. The behaviour of single modules, in this case the ASR, can impact the whole interaction. The robot being unable to understand the user's name does not necessarily seem like a very impactful problem at first, but it clearly shows that the overall system is not inclusive. Especially if the robot is collecting data from the user and is not just chit-chatting, it is important that the data is correct and that the user is able to provide the robot with the correct information.

\subsubsection{Takeaways}
These vignettes point to two crucial challenges with dialogue - saying relevant things and hearing each other (see Section~\ref{sec:functional-perspectives-hhi}). The LLM hallucinations described in \ztitleref{vignette:hallucinating} break \citeauthor{grice1975logicandconversation}'s maxim of \textit{quality}, where one should not say what is not supported by evidence \citep{grice1975logicandconversation} -- although it is unclear if an LLM can be said to hold a belief \citep{miehling2024languagemodelsdialogueconversational}.

\subsection{Challenges while conducting an experiment}
Even if significant effort is spent on resolving interaction problems during the development of a robotic system, issues can arise at a completely different scale when users actually get to interact with it -- in research, typically as it is used to run an experiment. We illustrate how vignettes can help to describe aspects that were generally challenging throughout the experiment - providing a way to report on what to improve that does not invalidate the study as such. We also show how vignettes can describe situations that would typically be dismissed as outliers, providing a way to discuss the larger challenges that they point to. 

\subsubsection{Vignette 3} \textit{Asking too long and complex questions:}
\zlabel{vignette:long-utterances}
We were conducting an experiment with elderly participants interacting with an attentive listening robot in a rest home. The participants talked with the robot about interesting experiences they had. One participant spoke about their overseas holiday for about 30 seconds. Then the robot responded by first acknowledging the information, explaining that they also liked the destination, and then asked a question about what they thought about the cultural landmarks there. The response lasted around 10 seconds. We expected the participant to immediately answer the question and continue speaking on that line of conversation. However the participant responded with ``Huh?'' and appeared confused. It appeared as if they could not catch what the robot was talking about. They proceeded with their talk on another subject, ignoring the question from the robot.

\paragraph*{Reflection} This vignette highlights that long turns, even ones meaningfully addressing the conversation, can be difficult to follow. This may be especially problematic for groups who may suffer more from hearing difficulties such as the elderly. While it is possible to modify an LLM prompt to mitigate this issue, this does not guarantee that the complexity of a question will be reduced. Instead of long and complex utterances, users require an empathetic response from the system, which could be very simple, such as asking for expansions (e.g. ``You drove a car?'') \cite{clark1996using}. 

\subsubsection{Vignette 4} \textit{Inability to give contextually relevant follow-up information:}
\zlabel{vignette:wine}
As part of an experiment that I conducted, a robot that we equipped with dialog capabilities was placed in the wine aisle of a supermarket to give customers advice on their wine selections. A poster informed customers about the robot's purpose of giving wine advice. The wine recommendations were based on information like the price, ingredients and the type of wine.
After customers had successfully interacted with the robot and received a wine recommendation, they were presented with the general information of the wine and a picture. However, the robot did not display any information on \textit{where} to find the wine on the shelves. When a customer asked the robot where they could find the recommended wine, it was unable to provide that information, apart from mentioning the general category of wine, which still left the customer with two shelves to search through by themselves. The customer voiced their frustration and did not start looking for the recommended wine. Instead of making use of the provided recommendation, they decided to choose a wine themselves, completely ignoring the previous interaction with the robot. 

\paragraph*{Reflection} This vignette highlights that not only the robot's appearance but also the specific context that it is placed in may shape users' expectations. The customers expect a robot capable of doing one thing (giving a recommendation) to also do the logical next thing (telling where to find the recommended item). If a human is capable of recommending a wine in a shop, then it is trivial for the human to also tell the customer where that wine is located in the shelf. As we saw in this vignette, this assumption does not hold for robots. System designers must be aware of what expectations users will have from a system based on its environment, placement and embodiment. LLMs may, in this case, give a more naturally expressed dialogue, but would not help find information about where in the aisle a specific wine is placed.

\subsubsection{Takeaways}
Turn design, embodiment and multimodality remain challenges for human-robot dialogue as shown by the vignettes. As described in Section~\ref{sec:functional-perspectives-hhi}, humans come to human-robot interactions under the assumption that those interactions will abide by basic communicative principles, whether those be Gricean \cite{grice1975logicandconversation} or grounding-based \cite{clark1996using, allwood1992feedback}. If the robot is not capable of aligning with the user's expectations, it falls on the user to adapt to the robot's style. In \ztitleref{vignette:long-utterances}, this would mean aligning with long utterances, causing poor interaction quality. In \ztitleref{vignette:wine}, there is no possible alignment between a customer who wants the robot to perform a task that it is incapable of performing.

\subsection{Challenges while demonstrating a system to the public}
When systems are deployed, many interesting interactions can be observed. In addition to more in-depth field studies, using vignettes to report on observations during demos can be a way to articulate persisting problems, which may point to weaknesses in the demonstrated system or serve as a motivation (or informal bug reports) for a new iteration of system development. 

\subsubsection{Vignette 5} \textit{Situationally inappropriate responses}

\zlabel{vignette:love-you}
I was invited to a panel discussion on artificial intelligence by the university, and one of the co-panelists brought a robot usually deployed in a local technology museum to demonstrate how LLMs can be used in robots. 
While preparing for the panel, I initiated a conversation with the robot, greeting it and asking what it could do. The setting was highly dynamic and noisy, so I was aware that the robot might not be able to handle the situation. While I expected that the robot might not respond, or answer that it had not heard what I said properly, I was instead surprised to find that the robot would generate a response in the noisy setting. When it responded, it became evident that the robot had not heard me correctly, as it responded ``I love you, too''. Even though I laughed off this unfitting statement in the moment, the scene came back to me when I later reflected on the event. The robot had done something that could be categorized as harassment if its programmer (who was standing right beside me) had said it in this setting. While mishearing and producing unfitting responses is common even in scripted robots, the fact that the response was generated by the LLM evoked new, unfamiliar feelings. Was the programmer to blame in this case at all? What happened in between my utterance and the robot's LLM that somehow considered this an adequate response?

\paragraph*{Reflection} This vignette illustrates that LLMs may present situationally inappropriate content, and it may be opaque to the user why this happened. While a scripted museum-robot might not say ``I love you'', an LLM can easily produce responses that violate the norms of the context where they are deployed. In the case of LLMs, it becomes utterly non-transparent who would be to blame -- is it the developer of the robot, the LLM, or is the robot reproducing stereotypes that came from its training data? One could set boundaries to prevent statements that might come off as inappropriate, but we hope to demonstrate with this example that a robot, which can more readily and proactively respond also has a much higher risk of generating utterances that are misfitting or experienced as potentially harassing, and that cannot be easily explained or recognized as an error in the automatic speech recognition.

\subsubsection{Vignette 6} \textit{Acting contextually inappropriate:}
\zlabel{vignette:hitting}
During a conference I attended, a humanoid robot with a display was placed in the foyer to provide information about the location, the conference and events. During a break, I decided to interact with the robot to get some information about the conference events that day.

After approaching the robot and being greeted by it, I asked for information about the conference. As a response, the robot displayed a long list of new options fitting the topic, while also providing a verbal answer. I decided to already have a look at the options while listening to the response. However, what I did not notice was that the robot was using gestures while talking. Since I was standing close to the robot and touching its tablet to scroll through the options, the robot hit me with one of its gestures. While the movement was not strong enough to hurt, it came as a surprise, since I had not been paying attention to the small random movements the robot had already performed earlier. After this, the robot just continued without acknowledging what just happened.

\paragraph*{Reflection} This vignette shows that spoken HRI is more than just speech -- it is also the combination of the speech with other modalities. The robot used speech to communicate with the user, but expected button presses on the tablet for selections. Even if those other modalities are used completely disconnected from the speech, it does not mean that they do not affect it. In the presented case, the complete disregard of hitting the user made the whole interaction feel surreal, since the robot had clearly broken a social norm, by first hitting me and then not acknowledging its inappropriate behaviour.

\subsubsection{Takeaways}
\ztitleref{vignette:love-you} and \ztitleref{vignette:hitting} illustrate the kinds of issues that arise when systems are demonstrated to the public. The kinds of systems that are demonstrated will by necessity be new and less developed than the more established systems that the prospective users are used to. This means that issues that had not been predicted by the designers and creators will arise -- like the ones seen in our vignettes. Such observations could stimulate future research directions, since they evoke more general questions about how interactions should be like. However, currently these issues are likely to go unreported, as it is hard to control the demo environment, record what happens (if the designers see the events to begin with), and manage participant consent. Formulating vignettes can be a way to capture the gist of what happened in a format that can stimulate discussion and reflection. 

\section{Discussion}
\label{sec:discussion}

Ethnographic vignettes can be both a way to report design problems and unforeseen issues when testing a system. They can serve as a way to report experiences that support design decisions for how to address common reoccurring problems. A multidisciplinary perspective is especially useful to figure out why a system that was built to fulfill a purpose does not do so, particularly due to complex failures of missing common ground \cite{clark1996using, allwood1992feedback} between the designer, user and system.

Semi-structured interviews help us access unstructured thoughts from participants without previously knowing what those thoughts are. In the same way, ethnographic vignettes can describe \textit{situations} that appear during interactions with our experiment for which we had not prepared.
Vignettes written from a development perspective enable us to share interesting observations which would otherwise not be part of the planned evaluation. They provide information in a descriptive way, accessible for researchers from other disciplines. 

It takes effort, time and publication space to present vignettes of interactions with a system. This needs to be weighed against how beneficial the vignettes are, but we argue that many vignettes like the ones we presented in Section~\ref{sec:vignettes} are less obvious than system designers believe that they are. Ethnographic vignettes should not replace quantitative analysis, but augment it with more nuanced observations. In a qualitative evaluation of a system, vignettes may provide more insights than aggregated user quotes from a survey.

Publishing vignettes in a paper with a strict page/word limit is challenging. They can end up filling up as much space as a full paper, but would generally not be considered to contribute equally much. We suggest placing them in supplemental materials to add context to the objectives of the study. Some publication venues have started introducing additional tracks where vignettes of system behaviour could be a great contribution on their own. Tracks like the video track at HRI or the case study track at CHI, can be suitable alternatives for reporting on practical experiences with HRI and HCI respectively. 

\section{Design suggestions}
\label{sec:design-suggestions}

Section~\ref{sec:discussion} leads us to two main study and workflow design suggestions for future work in HRI.
While recent proposals have pushed for higher standards of reporting methods, results, exclusion criteria and recruitment procedures in HRI \cite{bagchi2023towardsimprovedreplicability}, we focus on improving study design and reporting to include things that are currently not reported in the first place.

\subsection{Transparency about system capabilities}
Human-human interaction (HHI) quality is evaluated in ways that depend on the task being performed, and can be framed as a co-operative activity in which utterances are as informative as needed. This works because for many interactions we can immediately make assumptions about the goal.

Deficiencies like the ones in our vignettes highlight the differences between HHI and HRI. We can blame poor HHI on a lack of common ground on some level -- a lack of shared knowledge, differences in personality or what has been observed \citep{clark1996using}. These assumptions from HHI do not hold for robots, for which a human cannot assume that a common ground even exists \cite{miehling2024languagemodelsdialogueconversational}. Where humans engaging in HHI have mechanisms to address and repair interaction trouble on the fly \cite{Jefferson_2018}, issues in HRI may instead stem from faulty speech recognition \cite{kennedy2017childspeechrecognition,rudaz2023mpcosin} or dialogue models \cite{honig2018understandingandresolvingfailures}.

Because HRI involves humans, interactions will always be unpredictable. This can stem from preconceived notions about how the robot should respond, its capabilities, and the user's role in relation to the robot \cite{turkle2007authenticity, clarkfischer2023depictions}. 
Our vignettes show that these expectations are often not fully accounted for, and there is no easy method to identify and handle them. The presentation of the robots, including appearance and the environment, heavily influences expectations about the interaction and the robot's capabilities. Therefore, we can use these as grounding mechanisms. 
Matching user expectations to the capabilities of the robot is a precondition of a good interaction, but may still lead to a poor interaction if the capabilities are very limited.

User expectations can be tapped via pre-experiment questionnaires, similar to demographic information. In a laboratory setting, we propose that research focus more on qualitative measures such as subject interviews, co-design and even informal discussions as a form of analysis on the same level as quantitative surveys. Through these we can better understand user expectations which are likely to change over time.

\subsection{Reporting unexpected behaviours}

Some HRI evaluation standards have been used across the literature as highlighted in Section \ref{sec:interaction-quality}. These are required in the field since they are validated measures that can be used for comparisons over different contexts. On the other hand, many quantitative metrics will not capture the interactions described in Section \ref{sec:vignettes} partially because they may not necessarily happen frequently. Such cases can be used to describe flaws in robot or interaction design that still need to be tackled in the further development of dialogue systems for HRI. We therefore call for a critical perspective that does not only present the benefits and successes in using novel technology such as LLMs, but also seriously discusses its challenges.

We found the approach of writing vignettes useful for talking about unexpected behaviours of systems that we experienced. Formulating and narrating a scenario enabled us to discuss more concretely why the interaction was perceived that way, and what the underlying cause may have been. We propose that the HRI and dialogue system communities could benefit from using this approach to present system malfunctions at greater scale than what is currently done. 

\section{Conclusions}
In the future, we would like interactions between humans and robots to be natural, effective, honest and clear about their purpose, goal and capabilities. These are not easily measured or evaluated facets of an interaction, but by reporting HRI studies as a study of the robots' behaviours, including its failures, rather than as promotional materials for a fully functional system, we can move towards those goals.

\section*{Acknowledgments}
The authors would like to thank Felix Gervits for providing important feedback on an earlier version of this article. We would also like to thank the designers of the systems mentioned in \ztitleref{vignette:love-you} and \ztitleref{vignette:hitting} for giving us permission to use their systems as examples in this paper. This research was (partially) funded by the \href{https://hybrid-intelligence-centre.nl}{Hybrid Intelligence Center}, a 10-year programme funded by the Dutch Ministry of Education, Culture and Science through the Netherlands Organisation for Scientific Research, grant number 024.004.022. This work was supported by JST Moonshot R\&D JPMJPS2011.

\bibliographystyle{plainnat}
\bibliography{workshop_2025}

\appendix

\subsection{Vignette-writing instructions}
\label{sec:appendix-vignette-writing}
We found the following steps helpful to guide us through the process of writing vignettes. 

\textit{General instructions:} Plan some time to get into the writing mood, and give yourself space to recall the event in detail. You can imagine that you are describing the event to a friend. How would you tell them about the event? What information would you give to introduce the topic and let your friend imagine themselves there? 

Creating ``scenes on a page'' through writing can be done in different ways \citep[p. 45]{emerson2011writing}. Some writers prefer to jot down some notes first and elaborate later, while others produce longer sections of text immediately. 

We tried to cover three key aspects in a vignette:

\paragraph{Setting the scene} Try to recall the context in which the observed instance occurred. Where did it happen? Who was present? In what role were you participating in the event? Describe the scene in such a way that one can imagine being there with you.
\paragraph{Describing details of the action sequence} This will be the core of your vignette and you now want to zoom in on a specific situation, selecting a single moment in time. Describe the specific person(s) who are interacting with the robot and try to recall the order of the events as detailed as you can. You can use direct quotation (she said ``how are you?'') when you are very sure that something was phrased in a particular way. You may also use reported speech (she asked me how I was) and paraphrasing (we exchanged some greetings). While the focus may be on dialogue here, you also want to describe in detail what people did, certain movements or facial expressions. If you describe what the robot does, try not to say ``the robot expressed it was happy'' but rather be more specific - the robot smiled, played a sound with rising pitch, etc. You can then use a more interpretative stance in the next step. 
\paragraph{Reflecting on what made this reportable} Usually this step involves talking about feelings. What makes this event stick out as interesting to discuss? What made it surprising, frustrating or otherwise memorable? For whom? This is where you come in with how you experienced what happened from your first-person perspective. Try to pinpoint the social norms that the robot was breaching in the situation, for the people present. Often, it can help to explicitly formulate what would be expected in this situation and to contrast it with what actually happened. What did you and/or the people present expect to happen? What happened instead and how did you and/or the people present react next?

Especially when revising your vignette, try to capture the order of events as they evolved (often you just need to move some sentences around). What happened first, and what next?
Keep the programmers’ perspective out of the vignette---try to describe in social terms what happened. If you want, you can possibly add a fourth point at the very end if you really want to explain what happened from the robot's perspective. In that section, you should then purely focus on the robot/designer perspective, keeping the perspective of the users and the developers separate \cite{suchman2007human}.

\end{document}